\documentclass[pmlr]{jmlr}

\usepackage{longtable}

\usepackage{booktabs}
\usepackage[load-configurations=version-1]{siunitx} 
\usepackage{algorithm}
\usepackage{algorithmic}
\usepackage{mathtools}
\DeclareMathOperator*{\argmax}{arg\,max}

\makeatletter
\def\set@curr@file#1{\def\@curr@file{#1}} 
\makeatother

\theorembodyfont{\upshape}
\theoremheaderfont{\scshape}
\theorempostheader{:}
\theoremsep{\newline}

\jmlrvolume{126}
\jmlryear{2021}
\jmlrworkshop{Machine Learning for Healthcare}

\title[Medical symptom recognition from patient text]{Medical symptom recognition from patient text: An active learning approach for long-tailed multilabel distributions}

\author{%
\Name{Ali Mottaghi} \Email{mottaghi@stanford.edu}\\
\addr Stanford University
\AND
\Name{Prathusha K Sarma} \Email{prathyusha@curai.com}\\
\addr Curai
\AND
\Name{Xavier Amatriain} \Email{xavier@curai.com}\\
\addr Curai
\AND
\Name{Serena Yeung} \Email{syyeung@stanford.edu}\\
\addr Stanford University
\AND
\Name{Anitha Kannan} \Email{anitha@curai.com}\\
\addr Curai
}

\begin{document}

\maketitle

\begin{abstract}
  We study the problem of medical symptoms recognition from patient text, for the purposes of gathering pertinent information from the patient (known as history-taking). A typical patient text is often descriptive of the symptoms the patient is experiencing and a single instance of such a text can be ‘labeled’ with multiple symptoms. This makes learning a medical symptoms recognizer challenging on account of i) the lack of availability of voluminous annotated data as well as ii) the large unknown universe of multiple symptoms that a single text can map to. Furthermore, patient text is often characterized by a long tail in the data (i.e., some labels/symptoms occur more frequently than  others for e.g "fever” vs “hematochezia”). In this paper, we introduce an active learning method that leverages underlying structure of a continually refined, learned latent space to select the most informative examples to label. This enables the selection of the most informative examples that progressively increases the coverage on the universe of symptoms via the learned model, despite the long tail in data distribution.
\end{abstract}

\section{Introduction}
\label{sec:intro}

COVID-19 has catalyzed rapid adoption of telehealth \cite{Wosik2020} leading to greater than 100\% increase in virtual urgent care visits and greater than 4000\% increase in  virtual non-urgent care visits \cite{telemedicine}. A typical virtual visit starts with the patient describing their reason for their encounter/ visit (RFE) \cite{who}. 
Then, based on the RFE, an automated  history taking algorithm gathers further information including details about the presenting symptoms, and finally the patient is directed to a chat (video or text) with the practitioners.

\begin{figure*}
    \centering
    {\includegraphics[width=0.9\linewidth]{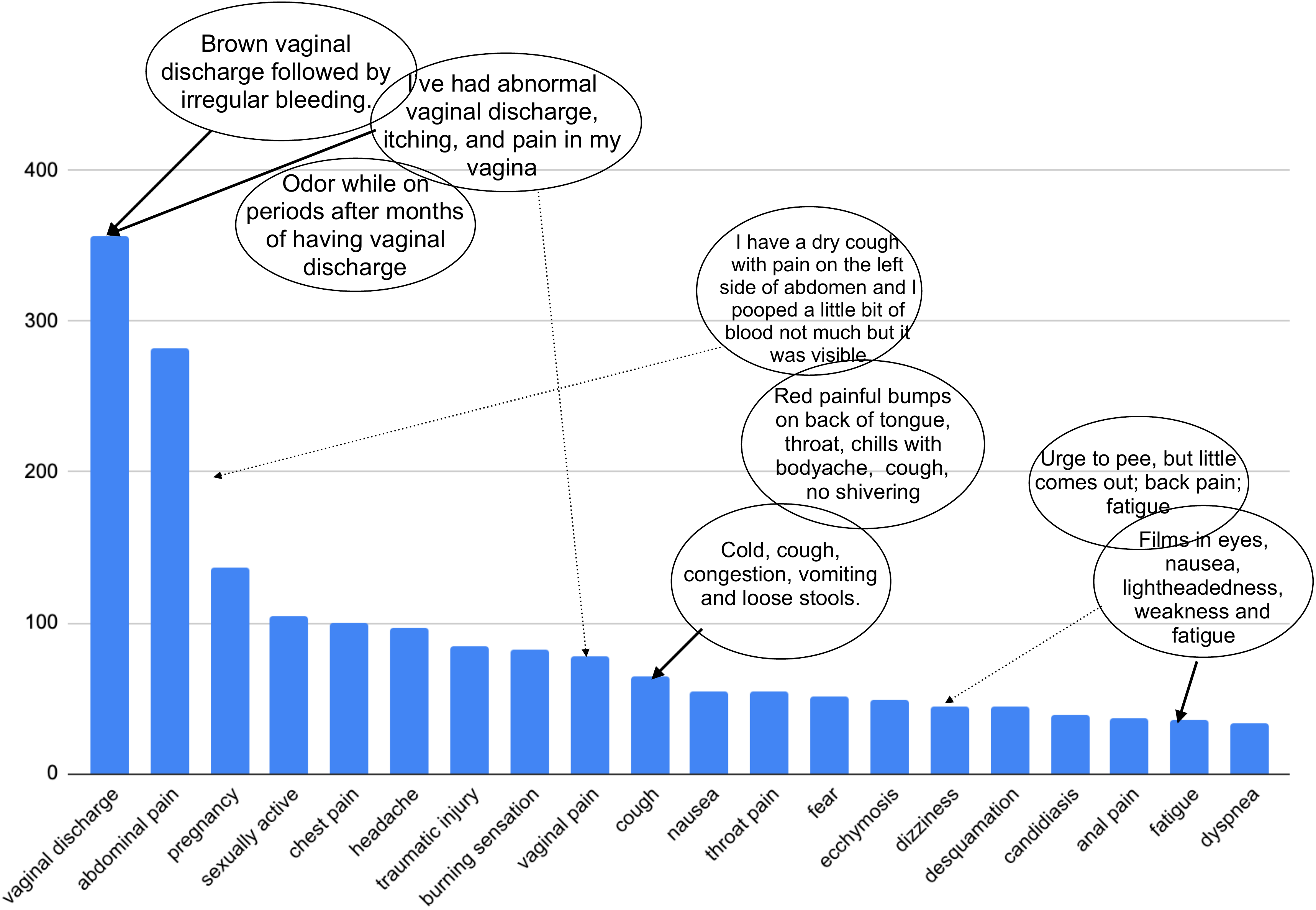}}
    \caption{
    Long-tailed data distribution. Also, shown example RFEs from dataset: Note the patient language and co-occurence of symptoms. 
    } 
    \label{fig:ml4hp}
\end{figure*}

This paper focuses on an intermediate implicit step of recognizing the symptoms in the patient's RFE, as it serves as the backbone for automated history taking. 
Note that this is related to the task of medical named entity recognition where the goal has been to extract symptoms
from the clinical text in electronic health records (c.f.\cite{Baumel18} and references therein). In contrast, we focus on medical entities recognition from the patient text;  because patient text (RFEs) are less standardized, and often express symptoms in a colloquial manner. Figure ~\ref{fig:ml4hp} presents some example RFEs from the dataset we use in our paper. 

Learning a model for medical symptom recognition poses a number of challenges: 
\begin{enumerate}
    \item \textbf{Access}: As there is no publicly available dataset, any new telehealth platform wanting to build its machine learning solution will need to start with a small set of labeled data points and rapidly increase the scope. This also means that as we increase the number of labeled data points, the system may need to recognize new symptoms that were previously unknown.
    \item \textbf{Multilabel distribution}: There may be multiple symptoms in a given RFE, and \emph{all} symptoms need to be extracted to enable correctly performing automated history taking. 
    \item \textbf{Long-tailed distribution}: Some symptoms are common while multiple symptoms tend to co-occur. This implicitly induces asymmetry and long tail in data distribution (Figure~\ref{fig:ml4hp}). This is compounded by the distribution of patient population on the telehealth platform.
\end{enumerate}

Given these challenges, in this work, we propose a new active learning algorithm for gathering data labels efficiently.  The main intuition for our algorithm is as follows: RFEs form distinct clusters that are medically grounded by underlying health conditions. Thus, the proposed algorithm selects informative examples by inducing clusterings in the latent space of unlabeled examples and choosing those that maximally cover that clustering. By finding the right balance between selecting points close to cluster centroids and points far from labeled RFEs, we can overcome the problem of long tail distribution. Our experiments show that our model outperforms the state-of-the-art on multi-label classification tasks and can address long-tailed distribution in our dataset. Ablation studies and qualitative analyses shed more light on the advantages of our approach.

\subsection*{Generalizable Insights about Machine Learning in the Context of Healthcare}
Deep learning has been applied to various applications in healthcare in recent years. One of the main characteristics of using deep learning for healthcare is dealing with the lack of large annotated data. It is usually costly and time-consuming to collect a large labeled dataset as annotations should be provided by healthcare professionals, and also datasets cannot be easily shared between institutions when working with sensitive and high-risk data. On the other side, deep models usually require large datasets to perform accurately and robustly. As a result, many AI applications in healthcare remain unreachable, and significant real-world impacts are delayed.

Most of the previous works on active learning focus on classification when the dataset is balanced. However, in many healthcare applications, the dataset has a long-tailed distribution, and the task is to find all the right labels (rather than the single right class). In the same vein, previous active learning strategies typically assume that only one label is valid per data point. In our paper, we show the importance of ``multilabel'' classification - a data point can have multiple correct labels. Equally importantly, for each data point,  correctly identifying  all labels and not missing any is of paramount importance for the subsequent task of history taking. In this work, we propose a new algorithm for actively selecting most informative examples to label, for the multi-label classification task. Our method utilizes the latent space of the machine learning model to overcome the long-tailed distribution of data and to select a diverse set of data points covering a wide range of inputs.

\section{Related Works}
\label{sec:related}

The main goal of active learning is to select a set of the most informative examples to be labeled and achieve the highest performance after training the model with the lowest number of data points being labeled. Active learning is a relatively established topic in machine learning, and most of the classic works in active learning can be found in \cite{settles2009active}. Active learning methods can be categorized into two main approaches: 1) \textbf{Pool-based methods} where examples are selected from a large pool of unlabeled data points. 2) \textbf{Query synthesizing methods} in which the active learning algorithm generates the data points for which it wants to get labels \cite{zhu2017generative} \cite{mahapatra2018efficient}. In this work, we focus on pool-based methods as they appear in many practical applications. 

There are three main approaches for pool-based active learning:
\noindent\textbf{Uncertainty approaches}: In these methods, an active learning algorithm selects the examples about which it is most uncertain, i.e., it labels data points that are most challenging for the model to predict. Different methods in this category measure the uncertainty in different ways such as entropy \cite{shannon1948mathematical}, variation \cite{freeman1965elementary} or standard deviation \cite{kendall2015bayesian} of output predictions or distance from the decision boundaries \cite{tong2001support}. \cite{gal2017deep} introduced a Bayesian approach called Bayesian Active Learning by Disagreement (BALD), which uses dropout in deep networks to capture the uncertainty. We compare to this line of work, but empirically we found this approach to to be not yet suited for our setting of long-tail multilabel classification. 

\noindent\textbf{Representation approaches}: In these approaches, the goal is to cover the latent space of the model and to select a diverse set of data points based on their latent representation. The Core-set \cite{sener2017active} method formulates the objective as a core-set selection in the latent space and selects a batch of examples with the highest coverage in the latent representation.  Our proposed approach falls under this line of work but extends to balance the trade-off between gathering more labeled examples for existing labels and identifying examples that covers the distribution of the latent space.

\noindent \textbf{Mixing Uncertainty and Representation}: Some approaches use both uncertainty and representation for selecting the data points. For example, \cite{yang2017suggestive} adds an uncertainty component to the core-set approach. \cite{ash2019deep} select data points in a hallucinated gradient space based on diversity and predictive uncertainty. Also, \cite{sinha2019variational} learns the distribution of labeled and unlabeled data points with a Variational Autoencoder (VAE), trains a discriminator for detecting the unlabeled data points based on latent representation, and sends the detected data points for annotation. This paper argues that we can learn both representation and uncertainty at the same time in this way. Our method utilizes the latent representation of the model for selecting a new data point, but it does not rely on the model's output given that we show the predictions are not reliable when the distribution is long-tailed, and some label categories are, yet, undiscovered. 

While most of the literature focuses on the classification task, a few works also tackle the problem of multi-label classification \cite{brinker2006active} \cite{yang2009effective} \cite{li2013active} \cite{vasisht2014active}.
Almost all of them use uncertainty as to the criteria for selecting data points. \cite{reyes2018effective} proposes a rank aggregation problem for measuring the uncertainty of each instance based on the predictions for all labels. \cite{nakano2020active} employs an ensemble of predictive models and uses the query-by-committee method to measure disagreement and uncertainty. We also further study this problem in the context of a long-tailed distribution dataset with undiscovered label categories which has not been studied before.

\section{Method}
\label{sec:method}

\begin{figure*}
    \centering
    {\includegraphics[width=0.9\linewidth]{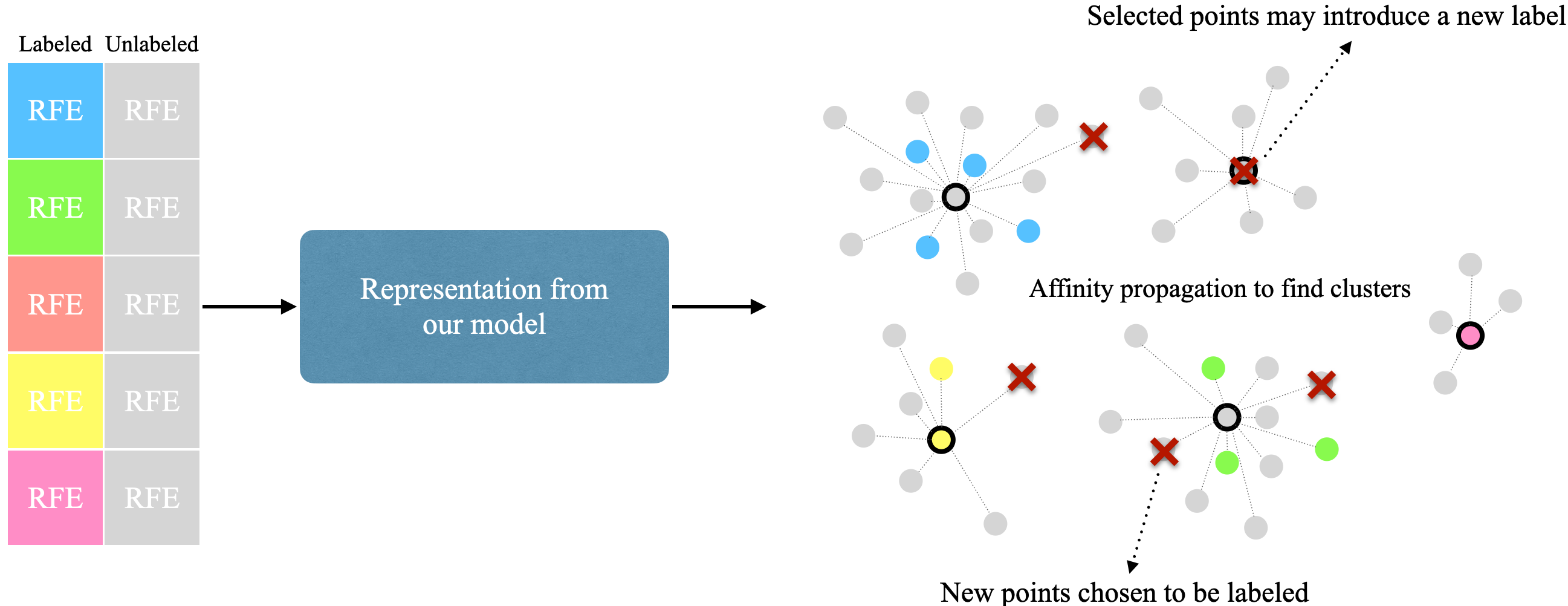}}
    \caption{
    Our active learning approach described in Alg.~\ref{alg}. In the beginning, only a small fraction of RFEs are labeled. The feature representation of all the RFEs is extracted using our model. Then Affinity propagation algorithm is utilized to reveal the underlying structure of the latent space. ALTL selects a diverse set of representations based on their distance to the cluster centroids and already labeled data points. Selected data points are labeled by an expert and are added to the labeled set. Selected data points might come from a new label that was not previously explored.
    } 
    \label{fig:method}
\end{figure*}

\paragraph{Notation:} Let $\mathcal{C}$ be the universe of symptoms.  Let ${x}= [w_1, w_2, \allowbreak \dots , w_t]$ represent a sequence of words corresponding to a reason for encounter (RFE), and  ${ y} \in \mathcal{Y} $ corresponds to a \emph{set} of labels. A labeled data point is represented by the pair (${x},{ y}$). Assume access to a labeled set $D_L = \{(x_1, y_1), (x_2, y_2), \dots , (x_n, y_n)\}$ and an unlabeled set $D_U = \{\bar{x}_1, \bar{x}_2, \dots , \bar{x}_m\}$. 

\paragraph{Algorithm:} Alg.~\ref{alg} provides an overview of Active Long-Tailed Learning (ALTL). At each iteration, the objective is to select $b$ unlabeled points from $D_U$ to be labeled by an oracle so as to minimize the future expected loss of the model after training on the new labeled set. We focus on batch active learning where $b > 1$ data points are selected at once since it is usually cheaper to label a batch of examples than labeling them one by one.

Unlike the typical active learning setup with balanced datasets, we consider the setup where the data distribution is long-tailed.  This mimics the real world setup wherein some symptoms (and hence labels) are much more common than others. Therefore, our dataset will correspond to a training set in which there is a variable number of RFEs corresponding to each label. This also means that our initial labeled set will not have samples corresponding to each symptom (and hence label), and more broadly, we may not have samples corresponding to all symptoms in the labeled set in some active learning iterations.  Our algorithm needs to be able to not only actively select the data points to expand the seen concepts but also to find undiscovered label categories. In the beginning of the model training, we have samples that correspond to a small subset of labels $\mathcal{C}$. In subsequent iterations of training, we want to select data points such that we obtain examples that expand existing labels and discover new ones, despite the data distribution being long-tailed. 
\\

{\noindent \textsc{Affinity}}: A standard approach is to choose a set of points that cover as much of the latent space of input representation as possible. While methods such as Core-set \cite{sener2017active}, BADGE \cite{ash2019deep}, and VAAL \cite{sinha2019variational} are tailored to this objective, they can also select data points that are outliers or are close to the boundaries of data representation; this issue is further exacerbated in our setting where the data distribution is long-tailed.  In order to address these issues, we leverage the Core-set framework but introduce several key innovations. First, we cluster the latent space of $D_L \cup D_U$ to capture the coherent structure.  This facilitates RFEs with different co-occurring symptoms to form distinct clusters (e.g. abdominal pain with a gynecological issue such as menstrual bleeding, vs. abdominal pain with constipation). For this, we use Affinity propagation \cite{frey2007clustering} as an exemplar clustering method based on probability propagation; it does not require the number of clusters to be specified beforehand.  This clustering step results in cluster centroids denoted as  $r_1, \dots, r_p \in \mathbb{R}^d$ where $p$ is the number of output clusters. 

\begin{algorithm}[t]
    \caption{Active Long-Tailed Learning (ALTL)} \label{alg}
    \begin{algorithmic}[1]
        \renewcommand{\algorithmicrequire}{\textbf{Input:}}
        \renewcommand{\algorithmicensure}{\textbf{Output:}}
        \REQUIRE Labeled set $D_L$, Unlabeled pool $D_U$, Budget $b$, Feature extractor $f(.)$
        \STATE Initialize $n = |D_L|$
        \STATE $r_1, \dots, r_p \gets \textsc{Affinity}(f(D_L \cup D_U))$
        \REPEAT 
        \STATE $u \gets \textsc{GetDatumToLabel}$($D_L$,$D_U$, $r_1, \dots, r_p, f(.)$)
        \STATE $y = \textsc{Oracle}(u)$
        \STATE$ D_L = D_L \cup \{(u, y)\}$
        \STATE$ D_U = D_U - \{u\}$
        \UNTIL $|D_L| = n + b$
        \RETURN $D_L$ and $D_U$
    \end{algorithmic} 
\end{algorithm}

\noindent{\textsc{GetDatumToLabel}} Then, we  select data points that explore both new regions of the latent space (and hence introduces previously unseen labels), as well as those that are close to the cluster centroids (captures data density in the latent space). We fold these two properties in our optimization function:
\begin{align}
    \label{acq}
    u &= \argmax_{x_u \in D_U} (\min_{x_l \in D_L \cup S}  \Delta(f(x_u), f(x_l)) - \ \lambda \min_{r \in \{r_1, \dots, r_p\}}  \Delta(f(x_u), r)) \nonumber
\end{align}

Here $\lambda$ is a hyper-parameter that balances the exploration of new parts of the latent space versus exploiting the established clusters, and $f(.)$ is the feature extractor. While increasing $\lambda$ enables selecting more examples near the cluster centroids where data points are concentrated, lowering it encourages exploring new areas in the latent space despite being far from established clusters in the data. As we show in our experiments, this approach is very effective when dealing with the long-tailed multilabel distribution of data. Figure~\ref{fig:method} shows an overview of our method.

\section{Experiments}
\label{sec:exp}

\paragraph{Data:} We gathered 1232 RFEs from an online telemedicine practice and had medical experts label them with corresponding symptoms. Table~\ref{tab:examples} provides example RFEs and their ground truth labels provided by the experts. RFEs are limited to 30-300 characters with an average length of around 100 characters. We consider a universe of 20 medical symptoms, where these 20 symptoms are made of the most frequently occurring medical findings in patient RFEs. We use 80:20 data split for train and test set, respectively. Data distribution is show in Figure~\ref{fig:ml4hp}. 

\begin{table*}
\footnotesize
\begin{tabular}{|lc|}
    \hline
    \multicolumn{1}{|c}{\textbf{Reason for encounter}}
    & \textbf{Symptoms} \\
    \hline
    \begin{tabular}{l}
        Left labia swelling, itching, bump
    \end{tabular} 
    & 
    \begin{tabular}{c}
        left labial edema\\ vaginal pruritus\\  vaginal nodule
    \end{tabular}
    \\
    \hline
    \begin{tabular}{l}
        I’ve had abnormal vaginal discharge, itching, and pain in my vagina
    \end{tabular} 
    & 
    \begin{tabular}{c}
        abnormal vaginal discharge\\ vaginal pain\\  vaginal pruritus
    \end{tabular}
    \\
    \hline
    \begin{tabular}{l}
        heavy bleeding, severe cramps and pain in left leg
    \end{tabular} 
    & 
    \begin{tabular}{c}
        heavy uterine bleeding\\ severe crampy abdominal pain\\ left lower extremity pain 
    \end{tabular}
    \\
    \hline
    \begin{tabular}{l}
        I have a dry cough with pain on the left side of abdomen and\\
        I pooped a little bit of blood not much but it was visible 
    \end{tabular} 
    & 
    \begin{tabular}{c}
        nonproductive cough\\ left abdominal pain\\ hematochezia
    \end{tabular}
    \\
    \hline
    \begin{tabular}{l}
        Urge to pee, but little comes out; back pain; fatigue.
    \end{tabular} 
    & 
    \begin{tabular}{c}
        urinary urgency\\ oliguria\\ back pain \\fatigue
    \end{tabular}
    \\
    \hline
\end{tabular}
\caption{Example Reasons for encounters, and the corresponding symptoms labeled by medical experts. In the first example, notice how itching and bump are associated with the anatomical part vagina because the patient discusses swelling of labia ("upper part of vagina"). Similarly, in the second example, notice the label `vaginal itching' based on the context. In the last example, "urge to pee, but little comes" is labeled to correspond to symptoms urinary urgency and oliguria.}
\label{tab:examples}
\end{table*}


\paragraph{Metrics:} We use F1-score~\cite{manning2008introduction} and Label Ranking Average Precision (LRAP)~\cite{tsoumakas2009mining} metrics for evaluating our model, which are widely used for multi-label classification tasks in machine learning.

\paragraph{Baselines:} We compare the performance of our method with the following baselines:
\begin{enumerate}
    \item \textbf{Random} baseline, as the name suggests, randomly selects new data points from the unlabeled data pool following a uniform distribution.
    \item \textbf{Fully-supervised} baseline is a fully supervised model trained on a completely labeled training data set, and serves as a reference. 
    \item \textbf{Core-set}~\cite{sener2017active}, as discussed in Section~\ref{sec:method} is a representation-based method that solves the  K-Center problem to select the points in the latent space. This baseline is the closest algorithm to our proposed method.
    \item \textbf{Max-entropy}~\cite{settles2009active} is the best known uncertainty-based method in Active Learning that uses the entropy of the model's output as a measure for uncertainty. Due to similarity in performance (also reported in \cite{sinha2019variational}) we do not compare against additional uncertainty-based baselines.
\end{enumerate}
In addition to these methods, we also compare against lookup based methods and Bayesian methods which we also report in \S~\ref{sec:main_results}

\paragraph{Model architecture:} Due to the size of available data sets as well as due to the well documented performance of pre-trained sentence encoders, we use InferSent~\cite{conneau2017supervised} for encoding patient RFE text. Each RFE is embedded in a $4096$ dimension vector. In addition to obtaining text embeddings from InferSent, a set of boolean feature vectors that capture the surface of symptoms from the patient text are created. These are $256$ dimensional embeddings where the presence of a 1 in the kth position indicates the presence of the kth symptom word (e.g. abdominal) in the patient text. Text embeddings are fed into a 2-layer MLP ($512$ and $256$ neurons) with $0.5$ dropout. Then the result is concatenated with boolean vectors. Another 2-layer MLP ($512$ and $128$ neurons) with $0.5$ dropout is used to generate the logits. The outputs before the last fully-connected layer are used as the main features in our method.

\paragraph{Training:} For training the model, we used the multi-label loss function proposed in \cite{mahajan2018exploring}. In this loss function, targets are formed by normalizing (sum to 1) the multi-hot encoded label vector, and then the cross-entropy of the target is computed with respect to the softmax of the model's output. For optimizing our loss function, we used Adam optimizer with $0.001$ learning rate. In Affinity propagation algorithm, we use euclidean distance and damping factor of $0.5$ for clustering. For all experiments, we train the model for $200$ epochs in each iteration. The results are averaged over $4$ runs, and predictions are made with $0.2$ margin on the softmax of the model's logits.  

\subsection{Main Results}
\label{sec:main_results}
To compare our proposed method with the baselines considered, we set up our experiments to first start with a small set of 10 labeled RFEs, and then at each iteration, we select a new batch (again of size 10) of unlabeled RFEs, via the acquisition function described in Alg.~\ref{alg}. The goal here is to choose the most representative RFEs to be labeled by an oracle. The new data points are then labeled by the oracle and added to the labeled training data set. The model is then retrained on this new training data set. This process is continued for multiple iterations until the labeling budget is exhausted.

\begin{figure*}[htbp]
  \centering
  \includegraphics[width=0.4\linewidth]{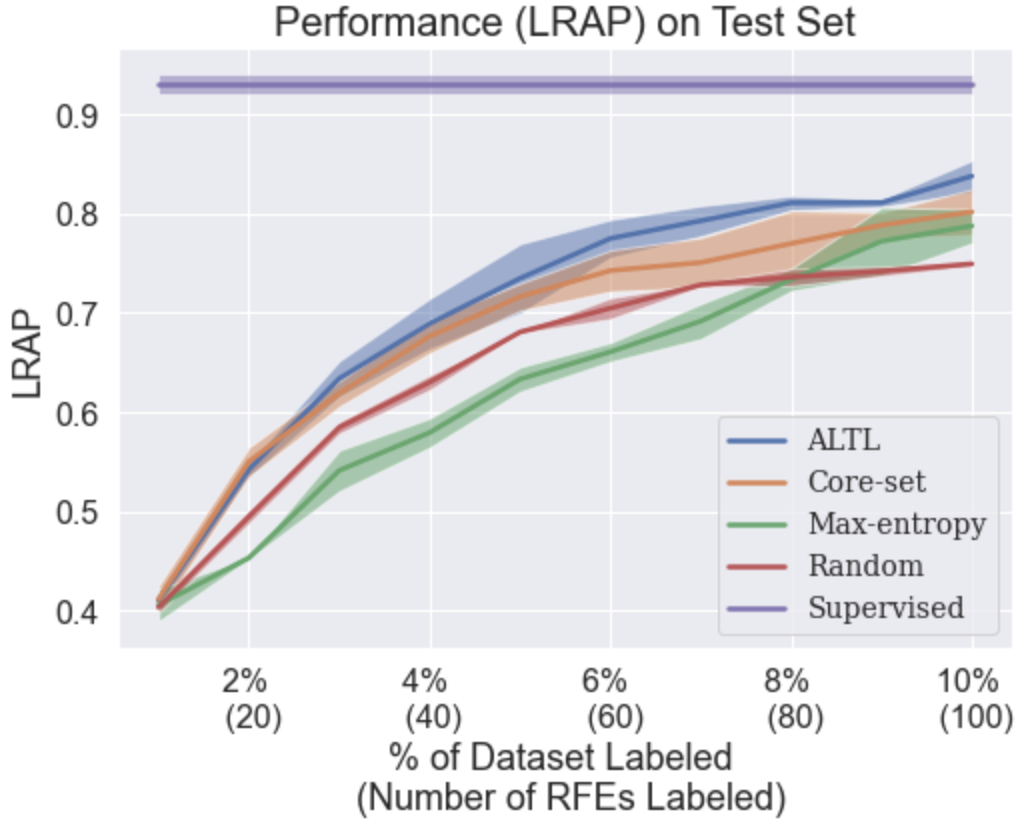}
  \hspace{1em}
  \includegraphics[width=0.4\linewidth]{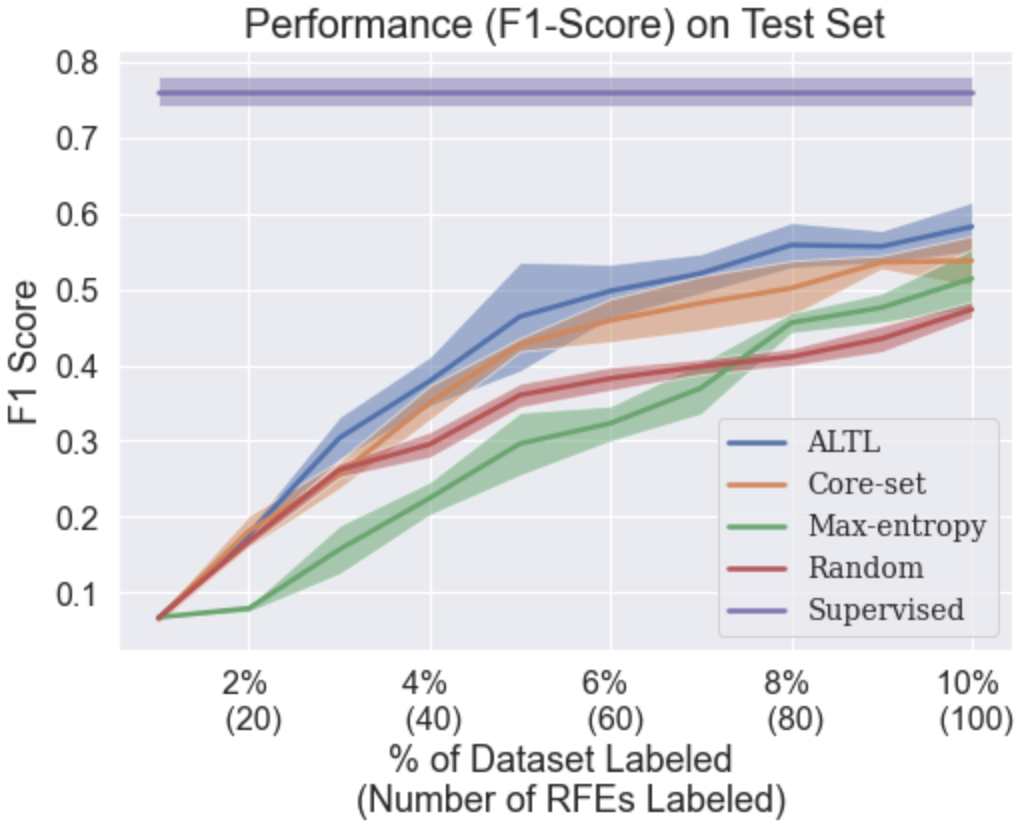}
  \caption{Performance (LRAP and F1-Score) of our method compared to Core-set, Max-entropy, Random, and Fully-supervised baselines.}
  \label{fig:baselines}
\end{figure*}

Figure~\ref{fig:baselines} shows the performance of our model compared to the baselines.  We can see that our approach outperforms the baselines. Since we start with only 10 labeled RFEs, many labels in the universe of symptoms do not have associated data samples. Our method is able to identify the previously unseen and unlabeled data points for exploration. 
Simultaneously, since the data has a long-tailed distribution, our methods are not susceptible to oversampling outliers along the boundaries of the data.
In contrast,  Core-set suffers from sampling too many outliers or data points too close to the boundaries in the latent space. Such examples are usually less informative than examples close to the centroids of the unexplored data clusters that our method can identify (see Qualitative Analysis~\ref{sec:qual} for clustering identified).  Max-entropy is not a competitive baseline since in the initial iterations, where many labels are still unexplored, the model's output is not a good indicator of uncertainty. Comparing against the fully supervised baseline, we see that our method achieves an LRAP score of 85 using only 10$\%$ of the labeled data set, as opposed to achieving a LRAP score of 92 when trained on the entire labeled data set. This shows that our method not only outperforms all the baselines, but also closes the gap with the fully supervised model when all the data points are labeled. The performance of the multi-label classification model is measured in both LRAP and F1-Score. LRAP focuses on the ranking of the output scores, while F1-Score takes into account the performance on each individual label (category). As it is shown in Fig. \ref{fig:baselines} our method performs similarly in both metrics. 

\paragraph{Comparison to lookup-based models:}
In order to validate the importance of a machine learned model, we also evaluated using a lookup based approach that uses a sliding window strategy to find maximal matches of text corresponding to the entities and their synonyms. Our model significantly outperforms this method as it better captures the semantics of the entire text. As an example, consider the RFE ``urge to pee, but little comes out". While the labels for this RFE are urinary urgency and difficulty urinating, one can imagine how a lookup based method would fail to identify the symptom `difficulty urinating'.  After being fully trained, our model achieves $76\%$ F1-Score, which is significantly higher compared to $58\%$ F1-Score for our best rule-based model.  

\paragraph{Comparison to Bayesian methods:} Many Bayesian approaches are also suggested for active learning, such as BALD \cite{gal2017deep} and BatchBALD \cite{kirsch2019batchbald}. However, these approaches usually rely on the fact that only one label exists in each sample, and they are suitable for classification only. As a result, they may perform poorly when used in a multi-label setting. We also applied BALD to our setting, but it performed significantly worse than other baselines, so we do not include that in the figure. With the same setting and after 10 iterations, BALD obtains $0.60 \pm 0.03$ LRAP and $0.25 \pm 0.03$ F1-Score, while our model achieves more than $0.80$ LRAP and around $0.60$ F1-Score (on average).

\begin{figure*}[htbp]
\floatconts
  {fig:subfigex}
  {\caption{ [best seen by zooming] TSNE visualisation of the latent space of the model and the selected data points by our method. The unlabeled points are shown as white dots and the labeled points are colored. The selected points by our method are marked. (a) At first iteration, our method chooses data points from a diverse set of clusters covering different parts of the latent space. (b) At last iteration, the points are selected such that they are not only far from the already labeled points, but also they capture undiscovered parts of the latent space.}}
  {%
    \subfigure[TSNE Visualisation of Features at First Iteration]{\label{fig:circle}%
      \includegraphics[width=0.75\linewidth]{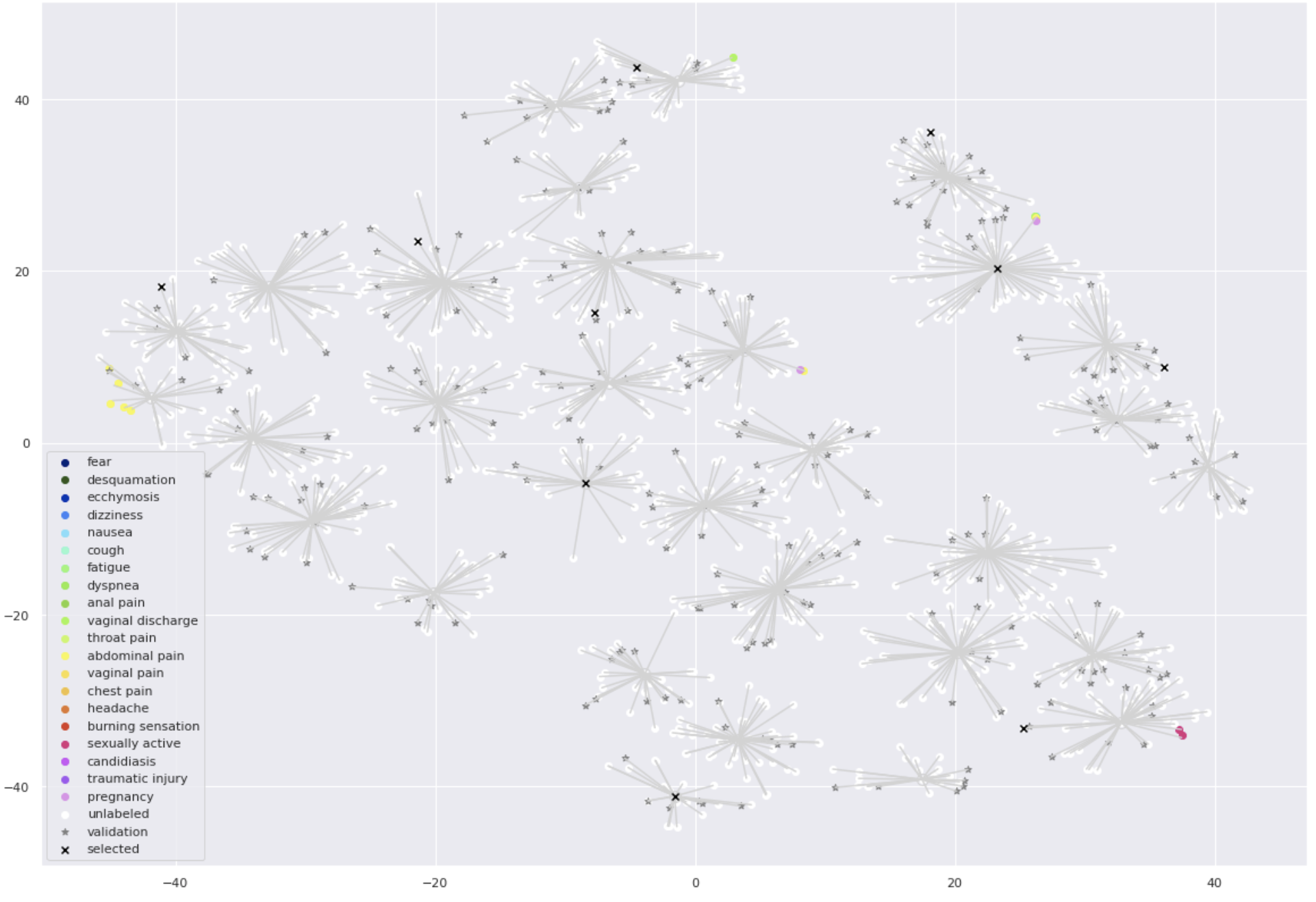}
      }%
    \qquad
    \subfigure[TSNE Visualisation of Features at Last Iteration]{\label{fig:square}%
      \includegraphics[width=0.75\linewidth]{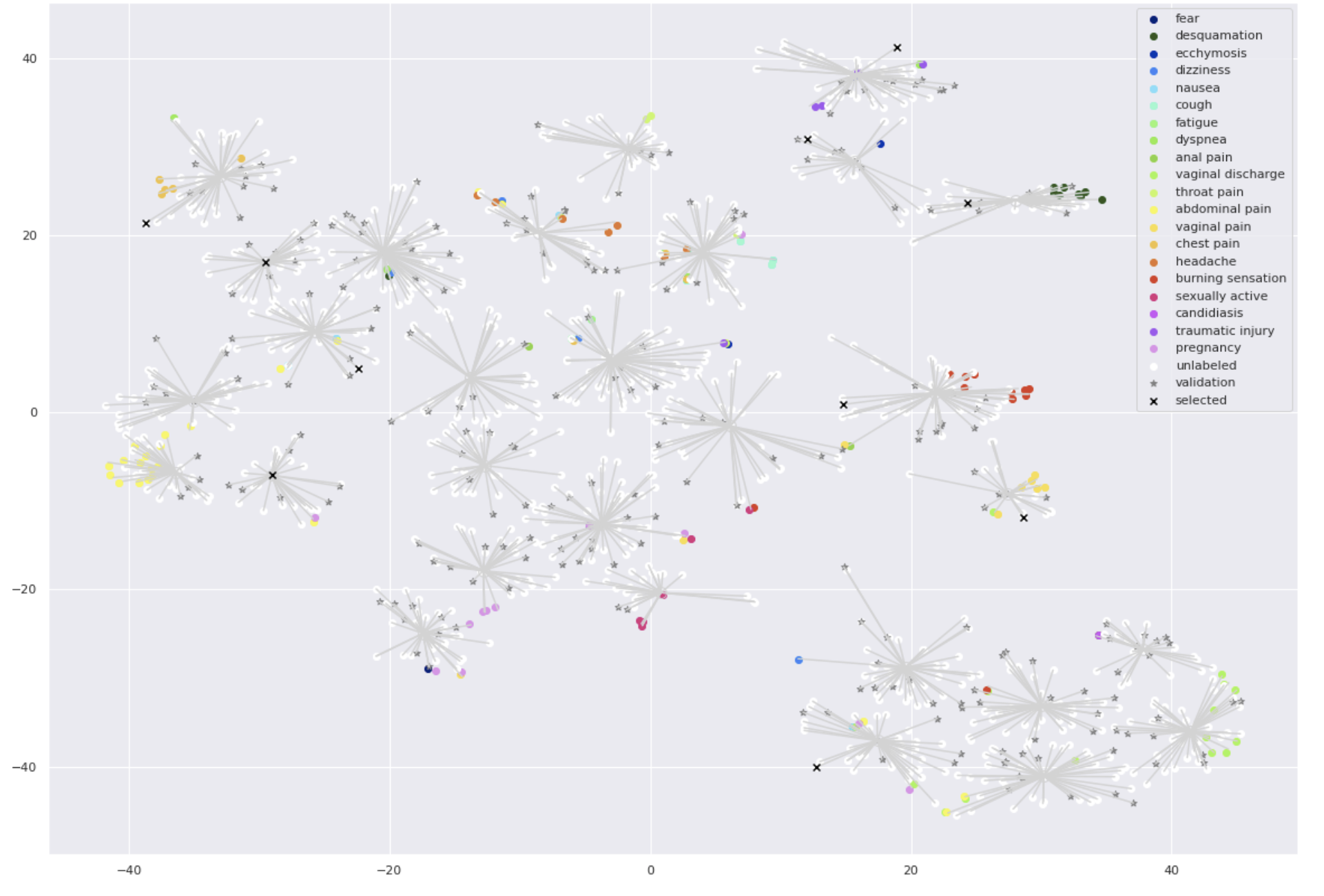}
      }
  }
  \label{fig:tsne}
\end{figure*}

\subsection{Ablation Study}
To understand the trade-off between exploring new clusters and exploiting the pre-existing clusters in the latent space, we perform the following ablation study. As described in Section~\ref{sec:method}, hyperparameter $\lambda$ is responsible for balancing the selection of points from the unexplored latent space vs. keeping points close to the centroid of clusters formed via Affinity Propagation.

Figure~\ref{fig:baselines} shows the difference in  performance as the function of $\lambda$. When $\lambda$ is meager, our method performs similarly to Core-set as expected since it does not account for the distance of the selected points from the cluster centroids. On the other hand, choosing a large $\lambda$ makes our method conservative in that it only selects cluster centroids. This is not helpful in the long-run, since cluster centroids of adjacent iterations are usually close to each other and as a result, selecting only centroids is not diverse enough; therefore, in this scenario, our method may under-perform Core-set. The optimal choice, in this case, is around $0.1$, which results in a good balance.

\begin{figure}[htbp]
  \centering
  \includegraphics[width=0.6\linewidth]{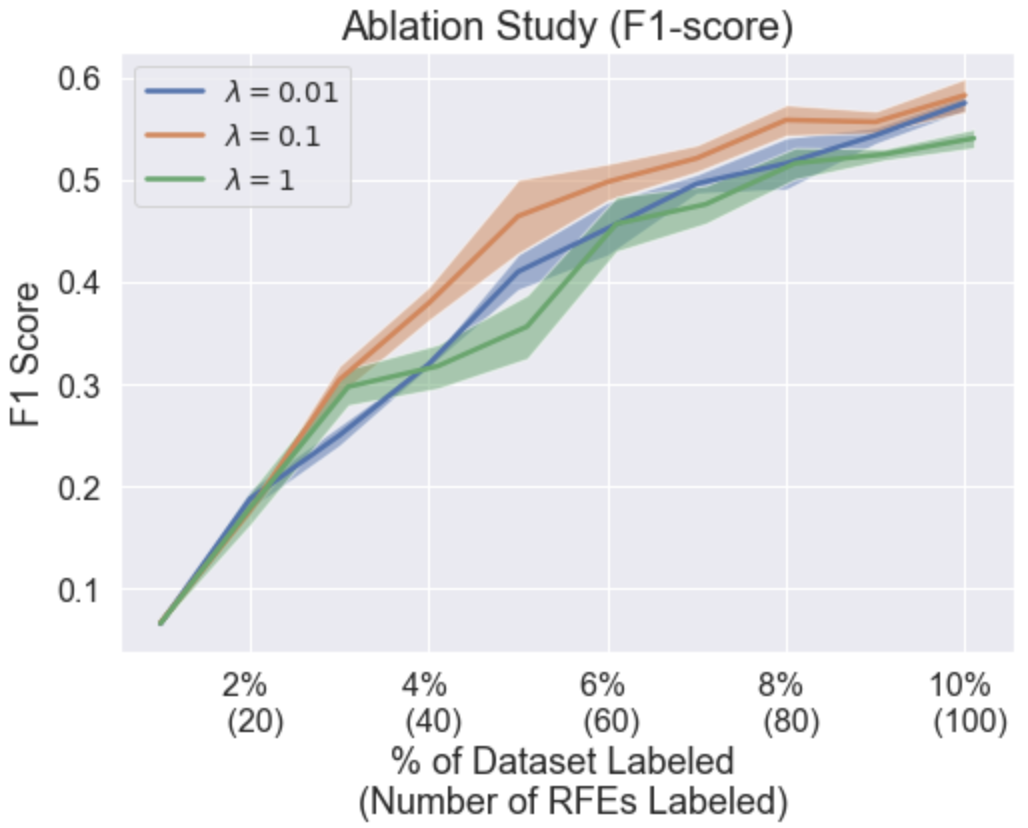}
  \caption{Ablation study of our method by varying the hyperparameter $\lambda$.}
  \label{fig:ab}
\end{figure}

\subsection{Qualitative Analysis}
\label{sec:qual}
We further analyze the behavior of our method qualitatively by visualizing the latent space. We visualize the features from the last layer of the fully connected network in our model using TSNE visualization~\cite{maaten2008visualizing}. We also include the clusters that resulted from the Affinity Propagation algorithm as disjoint graphs. Selected, labeled, and unlabeled data points are also depicted using crosses and colored and white dots in our graph. Figure~\ref{fig:tsne} shows the TSNE visualization of the latent space of the model in the first iteration and subsequent iterations as more data points are labeled. In the first iteration, our method selects a diverse set of data points in order to cover as much of the latent space as possible. However, it does not over-sample outlier data points or data points close to the boundaries, although they are attractive choices for the Core-set method. When more data points are already labeled in later interactions, our method continues to discover new clusters and improve upon the existing ones. This experiment shows that our model's qualitative behaviors are consistent with our hypothesis and expectations.

\section{Discussion} 
In this paper, we studied the problem of medical symptoms recognition from patient text. As we discussed in the paper, this problem can be formulated as a multi-label text classification problem. However, unlike most of the related works in the literature, the distribution of the data is long-tailed. We proposed a new active learning method that
can address long-tailed distribution in the dataset by selecting a diverse set of data points based on their latent representation. We use the affinity propagation clusters to better guide the selection of data points in the latent space. Our main goal is to take into account the density of data points in the latent space in addition to distance from labeled points. Clusters capture data points concentration. Therefore, selecting data points close to the cluster centroids usually results in covering more data points. However, simultaneously we want to explore undiscovered areas in the latent space to explore new labels. We propose a new active learning strategy which balances these two objectives, and selects data points based on their distance from the labeled points and cluster centroids. As we showed in our experiments, this can address long-tailed distribution in the dataset and outperform compared baselines in the symptoms recognition task.

\paragraph{Limitations} This paper presents our new active learning method for medical symptom recognition, as it serves as an excellent case study of a task with multi-label long-tailed distribution. However, long-tailed distributions are widespread in healthcare applications and appear in a wide spectrum of tasks. In our future works, we plan to study this problem for even more extreme long-tailed distributions. For example, in medical diagnosis, i.e., finding the underlying condition given a set of symptoms, the underlying conditions can cover a wide range of diseases. Another limitation is the data that support the findings of this work which is collected from an online telemedicine practice and is used under license for the current study. Therefore the data is not publicly available; however, the full code is available from the authors upon request.

\bibliography{jmlr}

\appendix

\end{document}